\documentclass{article} 
\usepackage{iclr2026_conference,times}

\usepackage{amsmath,amsfonts,bm}

\def\eqref#1{equation~\ref{#1}}

\def\1{\bm{1}}

\DeclareMathAlphabet{\mathsfit}{\encodingdefault}{\sfdefault}{m}{sl}
\SetMathAlphabet{\mathsfit}{bold}{\encodingdefault}{\sfdefault}{bx}{n}

\usepackage{hyperref}
\usepackage{url}

\usepackage{colortbl}
\definecolor{mygray}{gray}{.9}
\usepackage{algorithm} 
\usepackage{algpseudocode}
\usepackage{multirow} 
\usepackage{amsmath} 
\usepackage{amssymb}
\usepackage{amsmath}
\usepackage{CJKutf8}
\usepackage{courier}

\usepackage{graphicx}
\usepackage{booktabs}
\usepackage{multirow} 
\usepackage{multicol}
\usepackage{subcaption}
\usepackage{longtable}
\usepackage{booktabs}
\usepackage[most]{tcolorbox} 
\usepackage{makecell}  
 \newsavebox\CBox
\def\textBF#1{\sbox\CBox{#1}\resizebox{\wd\CBox}{\ht\CBox}{\textbf{#1}}}
\usepackage{wrapfig}
\title{On-the-Fly Data Augmentation via Gradient-Guided and Sample-Aware Influence Estimation}
\iclrfinalcopy

\author{
    Suorong Yang\textsuperscript{1 \dag}, Jie Zong\textsuperscript{1}, Lihang Wang\textsuperscript{1}, Ziheng Qin\textsuperscript{2}, \\
    \textbf{Hai Gan}\textsuperscript{1}, \textbf{Pengfei Zhou}\textsuperscript{2}, \textbf{Kai Wang}\textsuperscript{2}, \textbf{Yang You}\textsuperscript{2}  \textbf{Furao Shen}\textsuperscript{1 \dag} \\
    \textsuperscript{1} National Key Laboratory for Novel Software Technology, Nanjing University \\
    \textsuperscript{2} National University of Singapore \\
}

\begin{document}

\maketitle

\begin{abstract}
 Data augmentation has been widely employed to improve the generalization of deep neural networks. Most existing methods apply fixed or random transformations.
However, we find that sample difficulty evolves along with the model's generalization capabilities in dynamic training environments.
As a result, applying uniform or stochastic augmentations, without accounting for such dynamics, can lead to a mismatch between augmented data and the model's evolving training needs, ultimately degrading training effectiveness.
To address this, we introduce SADA, a Sample-Aware Dynamic Augmentation that performs on-the-fly adjustment of augmentation strengths based on each sample's evolving influence on model optimization.
Specifically, we estimate each sample’s influence by projecting its gradient onto the accumulated model update direction and computing the temporal variance within a local training window.
Samples with low variance, indicating stable and consistent influence, are augmented more strongly to emphasize diversity, while unstable samples receive milder transformations to preserve semantic fidelity and stabilize learning.
Our method is lightweight, which does not require auxiliary models or policy tuning. It can be seamlessly integrated into existing training pipelines as a plug-and-play module.
Experiments across various benchmark datasets and model architectures show consistent improvements of SADA, including +7.3\% on fine-grained tasks and +4.3\% on long-tailed datasets, highlighting the method's effectiveness and practicality.
\end{abstract}

\section{Introduction}
\label{sec:intro}
Data augmentation has been widely adopted for improving the generalization performance of deep neural networks~\citep{survey,survey2,survey3}.
Despite its effectiveness, most existing DA approaches remain static, non-adaptive, and sample-agnostic: they apply either fixed or randomly sampled transformations to all data uniformly, regardless of the evolving difficulty of individual samples or the dynamic learning state of the model in a dynamic training environment~\citep{trivialaugment,autoaugment,randaugment, dada}. 
For instance, methods such as Cutout~\citep{cutout}, AdvMask~\citep{advmask}, and Mixup~\citep{mixup} generate diverse training data by randomly sampling augmentation parameters.
Automatic methods, such as AutoAugment~\citep{autoaugment}, RandAugment~\citep{randaugment}, and DeepAA~\citep{deepaa}, search for dataset-specific augmentation policy space before training begins and then apply these fixed policies during training.
However, this design overlooks a crucial aspect of deep model training: the optimization landscape and the difficulty of individual samples evolve in dynamic training environments.
Some samples become easy to fit early on and require increased diversity to avoid redundancy, while others remain hard or unstable and should be preserved in their semantic form to support model refinement. Applying uniform augmentations across these heterogeneous cases introduces a mismatch between augmentation strength and training needs, potentially resulting in noisy updates, degraded sample utility, and suboptimal convergence.
Furthermore, many methods often require manual policy tuning or dataset-specific search, which limits scalability across different datasets and architectures~\citep{autoaugment,randaugment,entaugment}.
Adaptive augmentation approaches have emerged, but they typically involve bi-level optimization~\citep{madaug}, auxiliary models~\citep{teachaugment,yang2025adaaugment}, or large search spaces~\citep{freeaugment}, significantly increasing training complexity and resource demand.
Thus, a pressing question emerges:
\textit{Can we develop an on-the-fly augmentation mechanism that dynamically adapts training data to a model’s evolving learning dynamics without sacrificing scalability or efficiency.}

\begin{figure}
    \centering
    \includegraphics[width=0.9\linewidth]{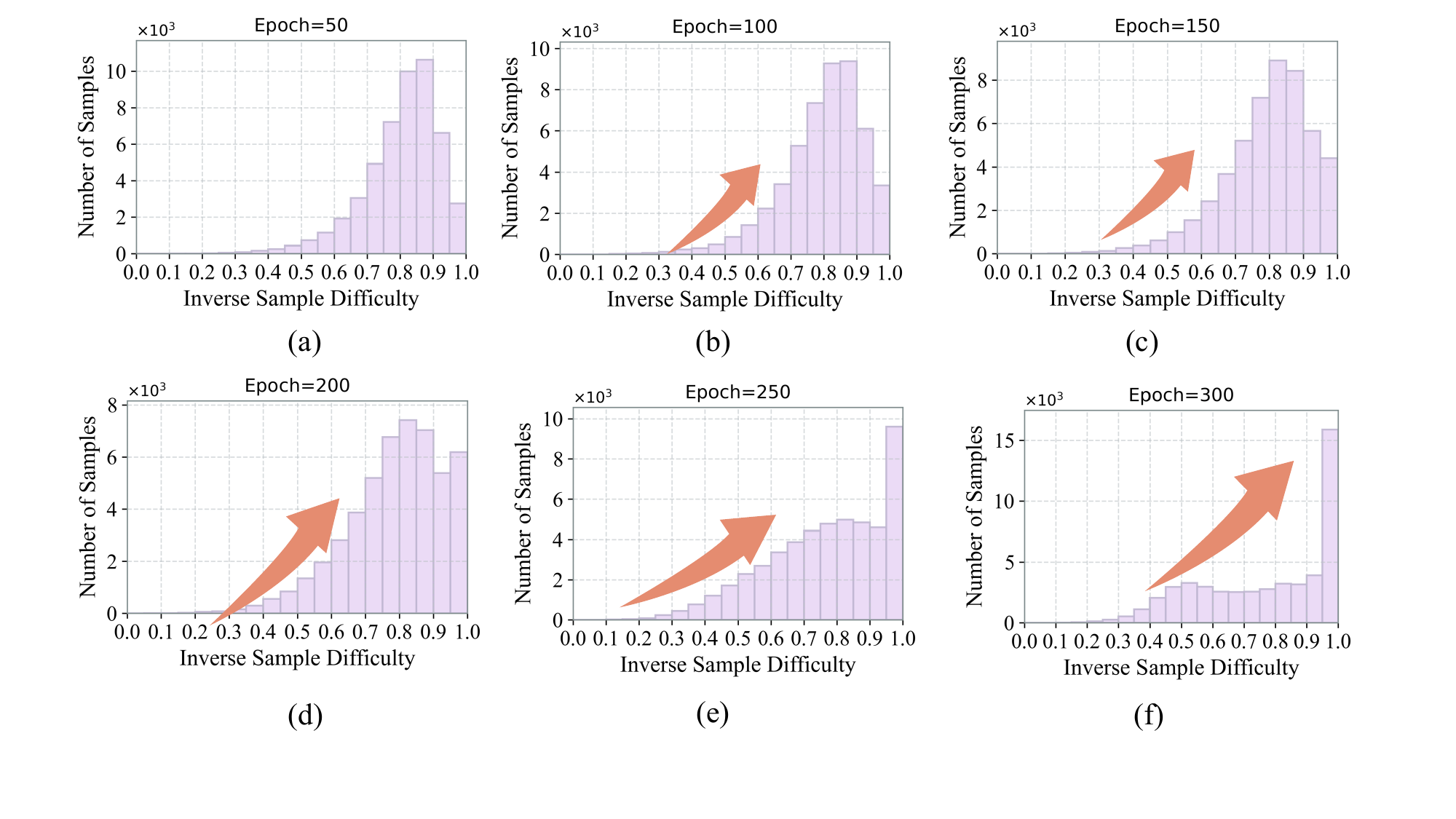}
    \caption{\textbf{Evolution of Sample Difficulty Across Training Epochs.} The distribution of sample difficulty evolves dynamically throughout training. Over time, a growing proportion of samples becomes easier (higher inverse sample difficulty values), particularly in later epochs, indicating a continuous shift in difficulty distribution during training. This dynamic trend highlights the necessity of dynamic and sample-aware augmentation strategies during training.}
    \label{fig1}
    \vspace{-5mm}
\end{figure}
In this paper, we propose SADA, a Sample-Aware Dynamic Augmentation method that performs on-the-fly adjustment of augmentation strength based on each sample’s evolving influence during training.
Unlike many existing methods that optimize augmentation operations~\citep{freeaugment,autoaugment}, our method uses a unified dataset- and model-agnostic augmentation space (refer to Table~\ref{tab:augmentation-space}) and directly modulates augmentation strength.
This design offers three benefits: 1). reducing the complexity of the decision space and ensuring efficient online training, 2). providing a more interpretable and fine-grained control over the trade-off between semantic consistency and diversity~\citep{investigating}, and 3). eliminating the need for manually crafted or optimization-required dataset-specific augmentation policies and enhancing scalability.
To quantify each sample's influence, we project its instantaneous gradient onto the direction of the accumulated model update, thereby capturing how much the sample contributes to the prevailing optimization trajectory.
The gradients can be naturally obtained during the standard forward and backward passes, ensuring high efficiency.
Furthermore, we compute the temporal variance of this projected influence within a local training window (e.g., 5 epochs), which serves as a proxy for the stability of a sample's learning dynamics.
When a sample exhibits consistently low variance, indicating a stable contribution to learning, more substantial augmentation is assigned to promote diversity and avoid overfitting to redundant patterns.
Conversely, samples with high variance, suggesting unstable or ambiguous influence, are augmented more conservatively to preserve semantic fidelity and support robust learning.
In this way, our method dynamically tailors augmentation magnitudes for each sample based on its training-stage-aware influence.
As illustrated in Figure~\ref{fig1}, our gradient-guided influence estimation reveals that sample difficulty continuously evolves throughout training: while more samples gradually become easier to fit as the model learns, a small subset remains persistently challenging.
By selectively increasing diversity for easier samples and preserving the core semantics of difficult ones, our framework improves generalization while mitigating the risk of introducing ambiguous or disruptive augmentations, highlighting the benefits of our sample-aware, dynamic augmentation. 

Experiment results across a variety of benchmark datasets and deep architectures demonstrate consistent and robust performance improvements. 
On benchmark datasets such as CIFAR-10/100~\citep{cifar}, Tiny-ImageNet~\citep{tiny}, and ImageNet-1k~\citep{imagenet}, our approach consistently outperforms existing data augmentation methods.
Additionally, we demonstrate strong generalization across different model architectures, including ResNet-based~\citep{resnet} and Vision Transformer (ViT)~\citep{vit}-based backbones, etc.
Thus, our method can be seamlessly integrated as a plug-and-play component without any modifications to model structures or training schedules.
On more challenging long-tailed datasets such as ImageNet-LT and Places-LT~\citep{lt-datasets}, models trained with our method achieve substantial gains, improving top-1 accuracy by over 4.3\% under the closed-set evaluation of ImageNet-LT, highlighting its robustness in imbalanced data scenarios.

Our main contributions can be summarized as follows: \textbf{(1)} We propose a lightweight, on-the-fly data augmentation framework that adjusts the augmented data based on the sample-aware evolving influence, without relying on auxiliary models or costly optimization procedures. \textbf{(2)} Our method explicitly captures the interplay between data and model by quantifying each sample's contribution to model optimization updates via gradient-guided influence estimation, aligning augmented data with the model's instantaneous learning dynamics. \textbf{(3)} Extensive experiments across diverse datasets and architectures demonstrate that our approach serves as a play-and-plug module, consistently improving generalization while maintaining training efficiency.

\section{Related Work}
\label{sec:related-work}
Data augmentation has long been a fundamental technique for mitigating overfitting and improving the generalization capability of deep neural networks. 
DA methods have evolved from simple, hand-crafted transformations to more adaptive and automated strategies.
It has evolved through multiple methodological paradigms. 
Early approaches primarily involved applying fundamental transformations, such as rotation, flipping, or cropping~\citep{Alexnet,survey}, to increase dataset diversity and model robustness. 
Subsequent advancements focus on developing more sophisticated transformation strategies tailored to specific data characteristics.
DA methods can be broadly categorized into image deletion-based, image fusion-based, and automatic policy-based  strategies~\citep{trivialaugment,entaugment}.

\textbf{Image Deletion-based Methods.}
Cutout~\citep{cutout} introduces regularization by randomly removing square regions from images.
GridMask~\citep{gridmask} generates resolution-matched masks for element-wise multiplication with images. 
Hide-and-Seek (HaS)~\citep{has} generalizes this idea by partitioning images into grids and stochastically masking subregions.
Random Erasing~\citep{randomerasing} further occludes rectangular areas without resizing. 
Moreover, AdvMask~\citep{advmask} generates learned or structure-aware masking to explicitly target semantic regions, encouraging the model to discover alternative discriminative cues.

\textbf{Image Fusion-based Methods.}
Fusion-based augmentation synthesizes training samples by blending information across multiple instances. 
Mixup~\citep{mixup} synthesizes samples via linear interpolation of pixel values and labels across image pairs.
However, its indiscriminate blending may produce visually incoherent samples. 
CutMix~\citep{cutmix} improves this by replacing rectangular regions between images, preserving spatial structure while introducing inter-sample variability. However, it may still obscure critical semantic content with irrelevant patches. 
Some improved variants, such as Attentive CutMix~\citep{attentivecutmix} and PuzzleMix~\citep{puzzlemix}, incorporate saliency awareness.
Despite their effectiveness, these methods typically rely on manually tuned parameters, with limited awareness of the model’s evolving training dynamics, potentially limiting the adaptability and optimization efficiency.

\textbf{Automated Augmentation Methods.}
Automated DA approaches define an augmentation operation space and search for optimal operations and magnitudes.
During training, the augmentation operation and corresponding magnitudes are randomly sampled from the pre-defined space.
AutoAugment (AA)~\citep{AA} uses reinforcement learning with an RNN controller to predict transformation sequences. 
Population-Based Augmentation (PBA)~\citep{pba} integrates genetic algorithms with parallel network training, while Fast AutoAugment~\citep{fast-autoaugment} employs Bayesian optimization to discover effective augmentation sequences across partitioned datasets. 
While powerful, these methods often incur high computational cost and are static once learned.
RandAugment~\citep{randaugment} and TrivialAugment~\citep{trivialaugment} simplify the parameter spaces through randomized policy selection. 
EntAugment~\citep{entaugment} uses entropy information derived from model snapshots to adjust the augmentation transformations, which may be affected by the inherent instability of model training.
Moreover, ParticleAugment~\citep{particleaugment} proposes a particle filtering scheme for the augmentation policy search.
Gradient-based DAS approaches formulate differentiable search spaces, enabling optimization of augmentation strategies. MADAO~\citep{MADAO} optimizes models and data augmentation policies simultaneously with Neumann series approximation of the gradients.
DADA~\citep{dada} formulates data augmentation policy search as a sampling problem and relaxes it into a differentiable framework via Gumbel-Softmax reparameterization. 
Adversarial variants such as Adversarial AutoAugment~\citep{Adversarialautoaugment} and TeachAugment~\citep{teachaugment} generate challenging transformations by maximizing training loss. DDAS~\citep{DDAS} exploits meta-learning with one-step gradient update and continuous relaxation to the expected training loss for efficient search, without relying on approximations like Gumbel Softmax.  In addition, DeepAA~\citep{deepaa} progressively constructs multi-layer augmentation pipelines.
FreeAugment~\citep{freeaugment} defines four free degrees of data augmentation and jointly optimizes them.
MADAug~\citep{madaug}, SelectAugment~\citep{selectaugment}, SLACK~\citep{slack}, and MetaAugment~\citep{metaaugment} optimize or learn sample-wise augmentation policies using various techniques, e.g., training an auxiliary policy network.
Despite these advances, most existing automated methods overlook the intrinsic heterogeneity of training data difficulty and fail to adapt augmentation intensities dynamically during online training. 
In contrast, our methodology introduces a lightweight, gradient-based mechanism that samples influence during training and adaptively adjusts augmentation magnitudes in real time, enabling fine-grained, instance-aware data augmentation.

\begin{figure}
    \centering
    \includegraphics[width=0.65\linewidth]{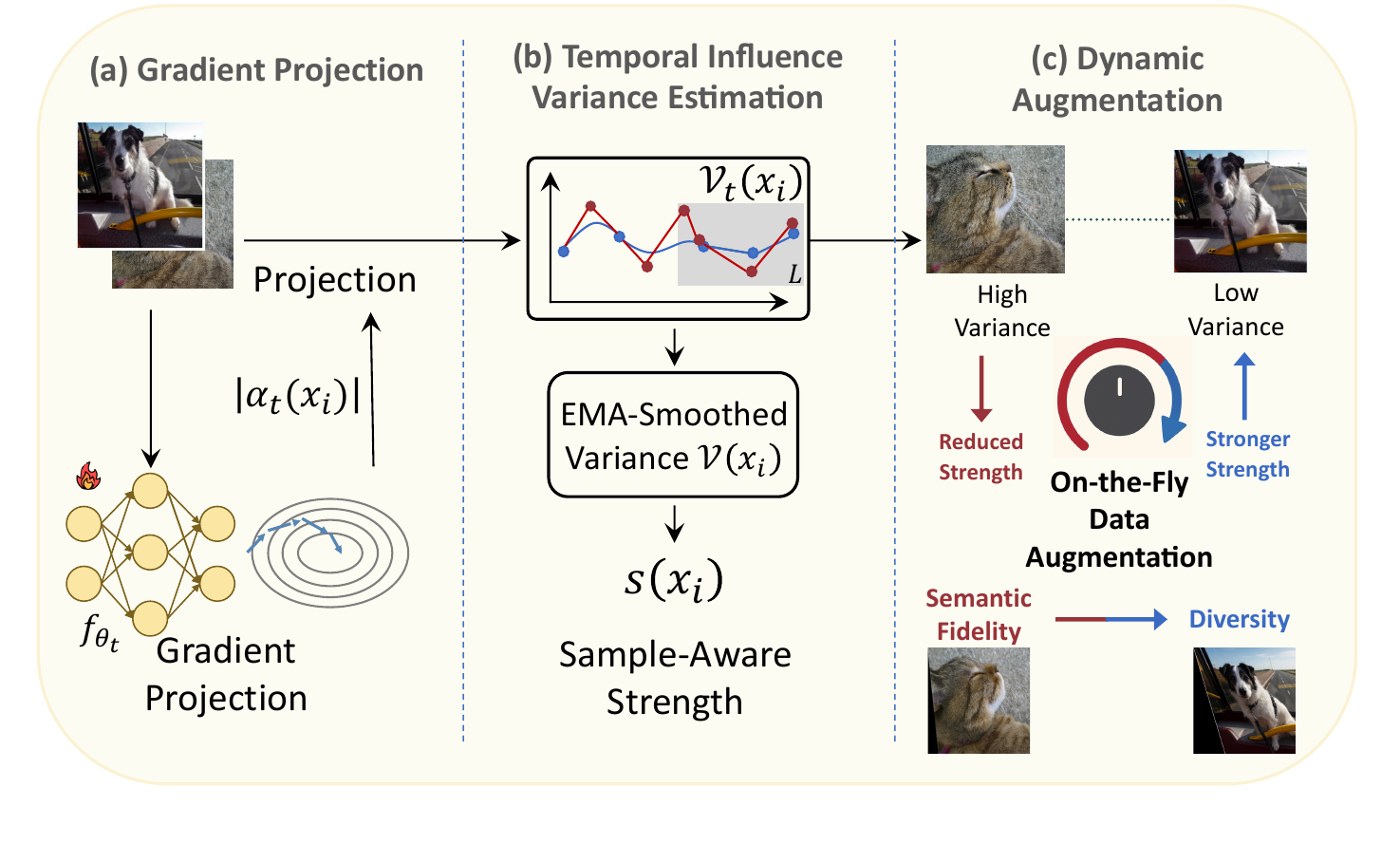}
    \caption{\textbf{Overview of Gradient-Guided On-the-Fly Data Augmentation.} At epoch $t$, we quantify the sample's influence on model optimization updates and estimate its stability. The augmentation strength is then adaptively adjusted based on this interplay between model training progress and sample difficulty.}
    \label{fig:framework}
    \vspace{-5mm}
\end{figure}

\section{Our Proposed Method}\label{sec:method}
\textbf{Overview.}
As illustrated in Fig.~\ref{fig:framework}, we propose an on-the-fly data augmentation method that adjusts sample-aware augmentation strength based on each sample's evolving influence on the model's optimization trajectory.
Specifically, we project the sample-wise gradient onto the accumulated gradient direction to quantify its contribution to parameter updates. 
To assess the consistency of this contribution, we compute the variance of the projected values within a local training window and apply EMA smoothing.
In this way, the augmentation strengths are dynamically determined in proportion to the stability of the sample's training influence.
Samples with low variance, indicating stable influence, are assigned stronger augmentations to improve generalization, while high-variance samples receive milder augmentations to maintain semantic fidelity and stabilize training.
In essence, our approach adjusts augmentation strength based on the interaction between the training data and the model's evolving optimization dynamics, thereby achieving dynamic augmentation.
During training, we randomly select one augmentation operation from the augmentation space for each sample per epoch and dynamically modulate its strength, which is uniformly applied to various datasets.

Let's denote the whole dataset as $\mathcal{D} = \{(x_i,y_i)\}^N_{i=1}$, where $x_i\in \mathbb{R}^D$, $y_i \in \mathbb{R}^{1\times K}$, and $K$ is the total number of classes.
The model $f_\theta$ is trained via gradient descent, updating parameters $\theta$ at step $t$ as:
\begin{equation} 
\theta_{t+1} = \theta_t -\eta \sum_{n=1,x_i\in \mathcal{D}}^N g_t(x_i),
\label{eq:update}
\end{equation}
where $g_t(x_i)$ is the gradient of the loss with respect to sample $(x_i,y_i)$, and $\eta$ is the learning rate.

During training, each sample contributes to the update of the model parameters via its gradient.
For samples that are easier to learn, the loss converges rapidly, and their gradient magnitudes tend to stabilize.
In contrast, more difficult and ambiguous samples often induce slower loss decay and exhibit persistently fluctuating gradients~\citep{forgetting,zhang2017understandingdeeplearningrequires,swayamdipta2020datasetcartographymappingdiagnosing}.
To quantify a sample's alignment with the model's current optimization trajectory, we compute the projection of its gradient onto the accumulated update direction.
Specifically, we focus on the projection value of the gradient in the direction of parameter updates.
The norm of the projected vector is calculated as follows:
\begin{equation}
|\alpha_t(x_i)| = |\left\langle g_t(x_i), \theta_{t-1}-\theta_t   \right\rangle|.
\end{equation}
The projected value reflects how much a sample's gradient contributes to the direction of the model’s parameter update.


To maintain high efficiency, we approximate the sample-wise gradient projection using first-order Taylor expansion, transforming the gradient-based formulation into a loss-based difference~\citep{tdds}.
Specifically, the projected influence $\alpha_t\left(x_i\right)$ can be approximated as:
\begin{equation}
    \begin{aligned}
\left|\alpha_t\left(x_i\right)\right| & =\frac{1}{\eta}\left|\left(\theta_{t-1}-\theta_{t}\right) \nabla_{\theta_{t-1}} \ell\left(f_{\theta_{t-1}}\left(x_i\right), y_i\right)\right| \\
& \approx \frac{1}{\eta}\left|\ell\left(f_{\theta_{t}}\left(x_i\right), y_i\right)-\ell\left(f_{\theta_{t-1}}\left(x_i\right), y_i\right)\right|,
\end{aligned}
\end{equation}
where $\ell(\cdot)$ denotes the loss function (e.g., cross-entropy).
This approximation reduces the need to compute inner products between gradients and parameter updates.
In the case of classification tasks with cross-entropy loss, the per-sample loss difference across consecutive steps is given by: 
\begin{equation}
\begin{aligned}
    \Delta \ell_{t-1}^n &= 
\ell(f_{\theta_{t}}(x_i),y_i) - \ell (f_{\theta_{t-1}}(x_i),y_i) \\
&= y_i^{\top} \cdot \log \frac{f_{\theta_{t}}(x_i)}{f_{\theta_{t-1}}(x_i)}.
\end{aligned}
\end{equation}
To generalize this formulation and enable a fully differentiable approximation, we replace the one-hot label with the soft target $f_{\theta_{t}}(x_i)^{\top}$, yielding a KL divergence between the model outputs at two consecutive steps:
\begin{equation}
\Delta \ell_{t-1}^n = f_{\theta_{t}}(x_i)^{\top} \cdot \log \frac{f_{\theta_{t}}(x_i)}{f_{\theta_{t-1}}(x_i)}.
\label{eq:gradient}  
\end{equation}
This formulation efficiently captures the alignment between a sample’s prediction dynamics and model update direction without computing explicit gradients.

To maintain high efficiency during training, we avoid complete historical gradient information and instead approximate sample influence using local training dynamics.
Specifically, we compute the variance of sample-wise loss differences over a fixed-size window of the past $L$ epochs, which is:
\begin{equation} 
\mathcal{V}_t(x_i) = \sum_{t-L+1}^t \left\| |\Delta \ell_t^n| - \overline{|\Delta \ell_t^n|} \right\|^2,
\label{eq:window-k} 
\end{equation}
where $\overline{|\Delta \ell_t^n|}$ denotes the average of the absolute loss differences within the window.
This formulation provides a local, memory-efficient measure of influence variability and mitigates instability from single-step snapshot assessments.
To smooth short-term fluctuations and emphasize recent training dynamics, we update the influence estimate using an exponential moving average:
\begin{equation}
\mathcal{V}(x_i) = \beta \mathcal{V}_t(x_i) + (1 - \beta) \mathcal{V}(x_i),
\label{eq:ema} 
\end{equation}
where $\beta$ is the decay coefficient and is set as a constant in our method.
In this way, the resulting influence score $\mathcal{V}(x_i)$ shows a proportional relationship with the sample difficulty.
To scale the values of $\mathcal{V}(x_i)$ into the range $[0,1]$, consistent with the allowable augmentation strength range $m_{max}$, we apply a min-max normalization on it and derive the applied augmentation strengths as $s(x_i)\cdot m_{max}$.
When $s(x_i) \rightarrow 1$, the augmented samples present a greater variability, and conversely, minor transformations occur as $s(x_i) \rightarrow 0$.
Importantly, $s(x_i)$ evolves dynamically throughout training, reflecting the model's changing perception of each sample's role in the optimization process.
Due to the limited space, we provide the details of the augmentation space and algorithm in Appendix~\ref{appendix:method-details}.

\noindent \textbf{Theoretical Analysis.} We provide a theoretical analysis to better understand why SADA works. 
In particular, we show that SADA reduces the empirical Rademacher complexity, thereby tightening the generalization error bound. Formally, the generalization gap is upper-bounded by a term of the form $\mathcal{O}(\tfrac{1}{n}\sqrt{\sum_i \alpha_i s_i^2})$, where $\alpha_i$ measures sample sensitivity to augmentation and $s_i$ denotes the applied augmentation strength.
Optimizing this bound yields a simple allocation rule: augment stable samples more, and unstable samples less. This aligns precisely with the SADA strategy.
Therefore, SADA improves generalization from data-centric learning.
The complete theoretical derivation is provided in Appendix~\ref{sec:theory}.

\noindent \textbf{Complexity Analysis.} 
We provide a theoretical analysis showing that SADA introduces negligible computational overhead compared to vanilla training.
Specifically, the computational complexity of our method is $\mathcal{O}( K \times N \times L)$, where $K$ is the total number of classes, $N$ is the number of samples, and $L$ denotes the window length. 
Since both $K$ and $L$ are constants and typically much smaller than $N$ (i.e., $L \ll N$, $K \ll N$).
As a result, the overall complexity simplifies to $\mathcal{O}(N)$, which is asymptotically equivalent to that of the CE loss used in the standard model training process.
Thus, our method incurs negligible additional training costs.


\vspace{-1mm}
\section{Experiment}

\begin{table*}[t]
\caption{Image classification accuracy (\%) on CIFAR-10\slash 100. * means results reported in the original paper~\citep{trivialaugment,entaugment}.}
    \label{tab:cifar}
	\centering
	\renewcommand\arraystretch{.9}
	\resizebox{0.97\textwidth}{!}{
			\begin{tabular}{c|llll|llll}
				\bottomrule[1.5pt]
   \multirow{2}{*}{Method} &  ResNet-44 &  ResNet-50
   & WRN-28-10 & SS-26-32 &   ResNet-44 & ResNet-50 
   & WRN-28-10 & SS-26-32 \\ \cline{2-9}
   &  \multicolumn{4}{c|}{ CIFAR-10 }& \multicolumn{4}{c}{ CIFAR-100 }\\ \hline
    baseline & 94.10\scriptsize{$\pm$.40} & 95.66\scriptsize{$\pm$.08}&95.52\scriptsize{$\pm$.11} & 94.90\scriptsize{$\pm$.07}*& 74.80\scriptsize{$\pm$.38}*&77.41\scriptsize{$\pm$.27}*&78.96\scriptsize{$\pm$.25}*&76.65\scriptsize{$\pm$.14}*\\    
    RE  &94.87\scriptsize{$\pm$.16}*&95.82\scriptsize{$\pm$.17}& 96.92\scriptsize{$\pm$.09}&96.46\scriptsize{$\pm$.13}* &75.71\scriptsize{$\pm$.25}*&77.79\scriptsize{$\pm$.32}&80.57\scriptsize{$\pm$.15}&77.30\scriptsize{$\pm$.18} \\
    RA& 94.38\scriptsize{$\pm$.22}&96.25\scriptsize{$\pm$.06}&96.94\scriptsize{$\pm$.13}*&97.05\scriptsize{$\pm$.15} &76.30\scriptsize{$\pm$.16}&80.95\scriptsize{$\pm$.22}&82.90\scriptsize{$\pm$.29}*&80.00\scriptsize{$\pm$.29} \\
    EA&  95.76\scriptsize{$\pm$.09}&97.09\scriptsize{$\pm$.09}&97.47\scriptsize{$\pm$.10}&97.46\scriptsize{$\pm$.11} &76.40\scriptsize{$\pm$.18}&81.56\scriptsize{$\pm$.21}&83.09\scriptsize{$\pm$.22} & 81.60\scriptsize{$\pm$.13} \\ 
    
    TA& 95.00\scriptsize{$\pm$.10}&97.13\scriptsize{$\pm$.08}&97.18\scriptsize{$\pm$.11}&97.30\scriptsize{$\pm$.10} &76.57\scriptsize{$\pm$.14}&81.34\scriptsize{$\pm$.18}&82.75\scriptsize{$\pm$.26}&82.14\scriptsize{$\pm$.16} \\
    AA& 95.01\scriptsize{$\pm$.11}&96.59\scriptsize{$\pm$.04}*&96.99\scriptsize{$\pm$.06}&97.30\scriptsize{$\pm$.11} &76.36\scriptsize{$\pm$.22}&81.34\scriptsize{$\pm$.29}&82.21\scriptsize{$\pm$.17}&81.19\scriptsize{$\pm$.19} \\
    FAA& 93.80\scriptsize{$\pm$.12}&96.69\scriptsize{$\pm$.16}&97.30\scriptsize{$\pm$.24}&96.42\scriptsize{$\pm$.12} &76.04\scriptsize{$\pm$.28}&79.08\scriptsize{$\pm$.12}&79.95\scriptsize{$\pm$.12}&81.39\scriptsize{$\pm$.16} \\
    HaS &94.97\scriptsize{$\pm$.27}&95.60\scriptsize{$\pm$.15}&96.94\scriptsize{$\pm$.08}&96.89\scriptsize{$\pm$.10}* &75.82\scriptsize{$\pm$.32}&78.76\scriptsize{$\pm$.24}&80.22\scriptsize{$\pm$.16}&76.89\scriptsize{$\pm$.33}  \\
     DADA& 93.96\scriptsize{$\pm$.38}&95.61\scriptsize{$\pm$.14}&97.30\scriptsize{$\pm$.13}*&97.30\scriptsize{$\pm$.14}*&74.37\scriptsize{$\pm$.47}&80.25\scriptsize{$\pm$.28}&82.50\scriptsize{$\pm$.26}*& 80.98\scriptsize{$\pm$.15}\\ 
    Cutout& 94.78\scriptsize{$\pm$.35}& 95.81\scriptsize{$\pm$.17}&96.92\scriptsize{$\pm$.09}&96.96\scriptsize{$\pm$.09}* &74.84\scriptsize{$\pm$.56}&78.62\scriptsize{$\pm$.25}&79.84\scriptsize{$\pm$.14}&77.37\scriptsize{$\pm$.28}  \\
    CutMix& 95.28\scriptsize{$\pm$.16}&96.81\scriptsize{$\pm$.10}*&96.93\scriptsize{$\pm$.10}*&96.47\scriptsize{$\pm$.07} &76.09\scriptsize{$\pm$.15}&81.24\scriptsize{$\pm$.14}&82.67\scriptsize{$\pm$.22}&79.57\scriptsize{$\pm$.10}\\ 
    GridMask &95.02\scriptsize{$\pm$.26}&96.15\scriptsize{$\pm$.19}& 96.92\scriptsize{$\pm$.09}&96.91\scriptsize{$\pm$.12} &76.07\scriptsize{$\pm$.18}&78.38\scriptsize{$\pm$.22}&80.40\scriptsize{$\pm$.20}& 77.28\scriptsize{$\pm$.38} \\
    AdvMask& 95.49\scriptsize{$\pm$.17}*&96.69\scriptsize{$\pm$.10}*&97.02\scriptsize{$\pm$.05}*&97.03\scriptsize{$\pm$.12}* &76.44\scriptsize{$\pm$.18}*&78.99\scriptsize{$\pm$.31}*&80.70\scriptsize{$\pm$.25}*& 79.96\scriptsize{$\pm$.27}* \\
    TeachA &95.05\scriptsize{$\pm$.21}&96.40\scriptsize{$\pm$.14}&97.50\scriptsize{$\pm$.16}&97.29\scriptsize{$\pm$.11} &76.18\scriptsize{$\pm$.31}&80.54\scriptsize{$\pm$.25}&82.81\scriptsize{$\pm$.26}&81.30\scriptsize{$\pm$.18} \\
    MADAug& 95.25\scriptsize{$\pm$.18}&97.12\scriptsize{$\pm$.17}&97.48\scriptsize{$\pm$.15}&97.37\scriptsize{$\pm$.11} &76.49\scriptsize{$\pm$.21}&81.40\scriptsize{$\pm$.18}&83.01\scriptsize{$\pm$.23}&81.67\scriptsize{$\pm$.19} \\
    SoftAug& 94.51\scriptsize{$\pm$.20}&96.99\scriptsize{$\pm$.14}&97.15\scriptsize{$\pm$.16}&97.22\scriptsize{$\pm$.19} &76.41\scriptsize{$\pm$.33}&80.94\scriptsize{$\pm$.33}&82.61\scriptsize{$\pm$.24}&80.33\scriptsize{$\pm$.20} \\  
     \hline
    Ours &  \textbf{95.87}\scriptsize{$\pm$.21}&\textbf{97.21}\scriptsize{$\pm$.10}&\textbf{97.66}\scriptsize{$\pm$.06}& \textbf{97.51}\scriptsize{$\pm$.07} &\textbf{80.81}\scriptsize{$\pm$.41}&\textbf{81.75}\scriptsize{$\pm$.28}&\textbf{83.17}\scriptsize{$\pm$.19}&\textbf{81.73}\scriptsize{$\pm$.15} \\

        \bottomrule[1.5pt]
		\end{tabular}}
        \vspace{-3mm}
    \end{table*}

\paragraph{Datasets and network architectures.}
Following prior works~\citep{trivialaugment,entaugment,autoaugment}, we evaluate our work on a diverse set of benchmark datasets, including CIFAR-10/100~\citep{cifar}, Tiny-ImageNet~\citep{tiny}, and ImageNet-1k~\citep{imagenet}.
To assess its effectiveness in fine-grained recognition tasks, we additionally conduct experiments on Oxford Flowers~\citep{oxford-flower}, Oxford-IIIT Pets~\citep{oxford-pets}, FGVC-Aircraft~\citep{oxford-aircraft}, and Stanford Cars~\citep{stanford-cars}.
Moreover, for evaluating performance under class imbalance, we also conduct experiments using long-tailed datasets, such as ImageNet-LT and Places-LT~\citep{lt-datasets}.
Due to the limited space, more experimental settings are provided in Appendix~\ref{app:settings}.

\paragraph{Comparison with State-of-the-arts.}
We compare our method with a wide range of representative and commonly used methods, including: 1). Cutout~\citep{cutout}, 2). HaS~\citep{has}, 3). CutMix~\citep{cutmix}, 4) GridMask~\citep{gridmask}, 5). AdvMask~\citep{advmask}, 6). RandomErasing~\citep{randomerasing}, 7). AutoAugment (AA)~\citep{autoaugment}, 8). Fast-AutoAugment (FAA)~\citep{fast-autoaugment}, 9). RandAugment (RA)~\citep{randaugment}, 10). DADA~\citep{dada}, 11). TeachAugment (TeachA)~\citep{teachaugment}, 12). MADAug~\citep{madaug}, 13). SoftAug~\citep{softaug}, and 14). TrivialAugment (TA)~\citep{trivialaugment}.



\subsection{Performance Comparison}
Table~\ref{tab:cifar} compares our method and several widely adopted state-of-the-art baselines on the CIFAR-10 and CIFAR-100 datasets across various deep architectures.
While the accuracy margins on these small-scale benchmarks are generally narrow, our method consistently achieves the highest performance across architectures.
For example, using WideResNet-28-10 on CIFAR-10, our approach improves accuracy by 2.14\% over the best-performing baseline. Similarly, with ResNet-44 on CIFAR-100, we observe a notable performance gain of 7.01\%.

To assess scalability, we further evaluate our method on the large-scale Tiny-ImageNet dataset in Table~\ref{tab:tiny}.
Across different architectures, our method consistently outperforms existing baselines.
For instance, on ResNeXt-50, it surpasses the next-best method by over 0.64\%, without introducing noticeable training overhead compared to standard training routines.
These gains can be attributed to our method's adaptive augmentation mechanism, which dynamically adjusts the augmentation strength based on each sample's influence stability.
This design enables a better balance between evolving models and training data, thereby enhancing generalization across models and datasets.

\subsection{Generalization on Large-scale ImageNet-1k}
We further evaluate the generalization performance of our method on the large-scale ImageNet-1k dataset. 
Specifically, following experiment settings~\citep{trivialaugment}, we train ResNet-50 models using different DA methods.
As shown in Table~\ref{tab:imagenet}, our method achieves a competitive performance compared to other baselines.
While the accuracy gap between our method and MADAug is marginal, our approach is significantly more efficient, achieving over 2x faster training than MADAug and over 4x faster than TeachAugment, without relying on auxiliary models or bi-level optimization.
These results demonstrate that our method offers a compelling trade-off between accuracy and efficiency for large-scale model training.
\begin{table}[]
 \caption{Image classification accuracy (\%) on Tiny-ImageNet across various deep models. }
		\label{tab:tiny}
	\centering
	\resizebox{0.7\columnwidth}{!}{
			\begin{tabular}{c|llll}
				\toprule[1.5pt]
				 Method & ResNet-18  & ResNet-50  & WRN-50-2 & ResNext-50  \\ \hline
     baseline &61.38\scriptsize{$\pm$0.99}&73.61\scriptsize{$\pm$0.43}&81.55\scriptsize{$\pm$1.24}& 79.76\scriptsize{$\pm$1.89}\\
     HaS &63.51\scriptsize{$\pm$0.58}&75.32\scriptsize{$\pm$0.59}&81.77\scriptsize{$\pm$1.16}& 80.52\scriptsize{$\pm$1.88}\\
     FAA  &68.15\scriptsize{$\pm$0.70}&75.11\scriptsize{$\pm$2.70}&82.90\scriptsize{$\pm$0.92}&81.04\scriptsize{$\pm$1.92} \\
     DADA &70.03\scriptsize{$\pm$0.10}&78.61\scriptsize{$\pm$0.34}&83.03\scriptsize{$\pm$0.18} & 81.15\scriptsize{$\pm$0.34} \\
     
    Cutout &68.67\scriptsize{$\pm$1.06}&77.45\scriptsize{$\pm$0.42}&82.27\scriptsize{$\pm$1.55}& 81.16\scriptsize{$\pm$0.78}\\
    CutMix  &64.09\scriptsize{$\pm$0.30}&76.41\scriptsize{$\pm$0.27}&82.32\scriptsize{$\pm$0.46}&81.31\scriptsize{$\pm$1.00} \\
    AdvMask  &65.29\scriptsize{$\pm$0.20}&78.84\scriptsize{$\pm$0.28}&82.87\scriptsize{$\pm$0.55}&81.38\scriptsize{$\pm$1.54} \\
    GridMask   &62.72\scriptsize{$\pm$0.91}&77.88\scriptsize{$\pm$2.50}&82.25\scriptsize{$\pm$1.47}& 81.05\scriptsize{$\pm$1.33}\\
    AutoAugment &67.28\scriptsize{$\pm$1.40}&75.29\scriptsize{$\pm$2.40}&79.99\scriptsize{$\pm$2.20}&81.28\scriptsize{$\pm$0.33} \\
    RandAugment   &65.67\scriptsize{$\pm$1.10}&75.87\scriptsize{$\pm$1.76}&82.25\scriptsize{$\pm$1.02}&80.36\scriptsize{$\pm$0.62} \\
    EntAugment &70.16\scriptsize{$\pm$1.01} & 79.06\scriptsize{$\pm$1.20} & 83.92\scriptsize{$\pm$0.97} & 81.90\scriptsize{$\pm$1.51} \\
    TeachAugment  &70.05\scriptsize{$\pm$0.57} &70.56\scriptsize{$\pm$0.44}&82.95\scriptsize{$\pm$0.13}&81.39\scriptsize{$\pm$0.97} \\
    TrivialAugment &69.97\scriptsize{$\pm$0.96}&78.41\scriptsize{$\pm$0.39}&82.16\scriptsize{$\pm$0.32} & 80.91\scriptsize{$\pm$2.26} \\
    RandomErasing &64.00\scriptsize{$\pm$0.37}&75.33\scriptsize{$\pm$1.58}&81.89\scriptsize{$\pm$1.40}& 81.52\scriptsize{$\pm$1.68}\\  \hline
Ours &\textbf{71.15}\scriptsize{$\pm$0.60}&\textbf{79.66}\scriptsize{$\pm$0.52}&\textbf{84.15}\scriptsize{$\pm$0.35}&\textbf{82.16} \scriptsize{$\pm$0.20} \\
    \bottomrule[1.5pt]
    \end{tabular}}
\end{table}

\begin{table*}[t]
     \centering
         \caption{Top-1 accuracy (\%) on ImageNet-1k dataset with ResNet-50.}\label{tab:imagenet}
	\renewcommand\arraystretch{1.2}
	\resizebox{1.\linewidth}{!}{
			\begin{tabular}{cccccccccccccccc}
				\toprule[1.pt]
				 HaS&GM&Cutout&CutMix& Mixup&AA&EA & FAA & RA &MA&SA&DADA&TA & TeachA 
      & Ours   \\ \hline
    77.2\scriptsize{$\pm$0.2}& 77.9\scriptsize{$\pm$0.2} & 77.1\scriptsize{$\pm$0.3} &  77.2\scriptsize{$\pm$0.2} & 77.0\scriptsize{$\pm$0.2}& 77.6\scriptsize{$\pm$0.2}&78.2\scriptsize{$\pm$0.2}&77.6\scriptsize{$\pm$0.2}&77.6\scriptsize{$\pm$0.2}&\textBF{78.5}\scriptsize{$\pm$0.1}&78.0\scriptsize{$\pm$0.1}&77.5\scriptsize{$\pm$0.1}&77.9\scriptsize{$\pm$0.3}&77.8\scriptsize{$\pm$0.2}
    & 78.4\scriptsize{$\pm$0.1}\\
    \bottomrule[1.pt]
    \end{tabular}}
        \vspace{-5mm}
\end{table*}
\begin{table*}[t]
     \centering
     \setlength{\tabcolsep}{2.5pt}
         \caption{Transferred test accuracy (\%) on CIFAR-10 of various DA methods. The pretrained ResNet-50 model is trained on CIFAR-100 (upper row) and Tiny-ImageNet (bottom row).}\label{tab:transfer-learning}
	\renewcommand\arraystretch{1.}
	\resizebox{.99\linewidth}{!}{
			\begin{tabular}{ccccccccccccccc}
				\toprule[1.pt]
    baseline &HaS & FAA & DADA & Cutout & CutMix & MADAug & GridMask & AA&EA & RA & TeachAug & TA &RE & Ours \\ \hline
 91.53\scriptsize{$\pm$.03}&92.51\scriptsize{$\pm$.24}&92.28\scriptsize{$\pm$.13}&92.58\scriptsize{$\pm$.09}&92.42\scriptsize{$\pm$.20}&92.81\scriptsize{$\pm$.47}&92.84\scriptsize{$\pm$.10}&91.49\scriptsize{$\pm$.10}&92.82\scriptsize{$\pm$.04}&92.89\scriptsize{$\pm$.19}&92.78\scriptsize{$\pm$.23}&92.83\scriptsize{$\pm$.18}&92.80\scriptsize{$\pm$.16}&92.55\scriptsize{$\pm$.05}&\textbf{93.11}\scriptsize{$\pm$.25}\\
 64.02\scriptsize{$\pm$.05}&66.84\scriptsize{$\pm$.06}&70.32\scriptsize{$\pm$.63}&69.04\scriptsize{$\pm$.43}&65.54\scriptsize{$\pm$.75}&69.29\scriptsize{$\pm$.09}&72.82\scriptsize{$\pm$.32}&64.88\scriptsize{$\pm$.43}&69.53\scriptsize{$\pm$.53}&72.68\scriptsize{$\pm$.73}&64.68\scriptsize{$\pm$.23}&69.98\scriptsize{$\pm$.17}&71.53\scriptsize{$\pm$.35}&64.56\scriptsize{$\pm$.27}&\textbf{77.26}\scriptsize{$\pm$.12}\\
    \bottomrule[1.pt]
    \end{tabular}}
\end{table*}
\begin{table*}[ht]
 \caption{Top-1 classification accuracy (\%) on ImageNet-LT and Places-LT. * means results reported in the original paper. }
		\label{tab:imagenet-LT}
    \centering
    \setlength{\tabcolsep}{2.5pt}
        \resizebox{.99\textwidth}{!}{\begin{tabular}{c|l|cccc|cccc}
          \bottomrule[1.1pt]
        \multirow{2}{*}{Dataset}& \multirow{2}{*}{Methods} &\multicolumn{4}{c}{ closed-set setting}&\multicolumn{4}{|c}{ open-set setting} \\ \cline{3-10}
        & & \textbf{Many-shot} & \textbf{Medium-shot} & \textbf{Few-shot} & \textbf{Overall}  & \textbf{Many-shot} & \textbf{Medium-shot} & \textbf{Few-shot} & \textbf{F-measure}\\ \hline
      \multirow{2}{*}{ImageNet-LT} & OLTR  & 43.2\scriptsize{$\pm$0.1}* & 35.1\scriptsize{$\pm$0.2}* & 18.5\scriptsize{$\pm$0.1}* & 35.6\scriptsize{$\pm$0.1}* &41.9\scriptsize{$\pm$0.1}*& 33.9\scriptsize{$\pm$0.1}*& 17.4\scriptsize{$\pm$0.2}* &44.6\scriptsize{$\pm$0.2}* \\ 
        &  OLTR+\textbf{Ours}&
        \textbf{46.9\scriptsize{$\pm$0.1}} &
        \textbf{37.0\scriptsize{$\pm$0.2} }&
        \textbf{21.6\scriptsize{$\pm$0.2}} &
        \textbf{36.9\scriptsize{$\pm$0.1}} &
        \textbf{45.2\scriptsize{$\pm$0.1}} &
        \textbf{35.6\scriptsize{$\pm$0.2}} &
        \textbf{20.6\scriptsize{$\pm$0.1}} &
        \textbf{45.5\scriptsize{$\pm$0.1}} \\ 
        \hline
      \multirow{2}{*}{Places-LT } &  OLTR  & \textbf{44.7}\scriptsize{$\pm$0.1}* & 37.0\scriptsize{$\pm$0.2}* & 25.3\scriptsize{$\pm$0.1}* & 35.9\scriptsize{$\pm$0.1}*  &\textbf{44.6}\scriptsize{$\pm$0.1}*& 36.8\scriptsize{$\pm$0.1}*& 25.2\scriptsize{$\pm$0.2}* & 46.4\scriptsize{$\pm$0.1}*\\
        &  OLTR+\textbf{Ours} & 
        44.3\scriptsize{$\pm$0.1} &
        \textbf{40.8\scriptsize{$\pm$0.2}} &
        \textbf{28.9\scriptsize{$\pm$0.2}} &
        \textbf{38.5\scriptsize{$\pm$0.1}} &
        44.1\scriptsize{$\pm$0.1} &
        \textbf{40.6\scriptsize{$\pm$0.2}} &
        \textbf{28.6\scriptsize{$\pm$0.1}} &
        \textbf{50.4\scriptsize{$\pm$0.2}} \\ 
         \bottomrule[1.1pt]
        \end{tabular}}
\end{table*}
\subsection{Data Augmentation Improves Transfer Learning}
Beyond evaluations on benchmark datasets, we assess model generalization through transfer learning, which tests a model's ability to extract transferable and robust features across domains~\citep{transfer-learning-1,transfer-learning-2,transfer-learning-3}.
In this setup, we pretrain ResNet-50 models on CIFAR-100 and Tiny-ImageNet using various data augmentation methods, and then fine-tune them on CIFAR-10.

This evaluation is motivated by the fact that stronger data augmentation strategies can lead to more generalizable feature representations.
As shown in Table~\ref{tab:transfer-learning}, it can be observed that our method achieves consistently higher accuracy after transfer compared to baseline augmentation approaches, regardless of the pertaining dataset.
These results indicate that models trained with our dynamic augmentation strategy learn more transferable and semantically meaningful features, further validating the generalization benefits of our approach.

\begin{figure*}[t]
\centering
    \begin{minipage}{0.3\textwidth}
     \captionof{table}{Test accuracy (\%) on fine-grained datasets with ResNet-50.}
	\renewcommand\arraystretch{1.}
	\resizebox{1\textwidth}{!}{
        \begin{tabular}{l|c c}\toprule[1.pt]
    Dataset & baseline & Ours \\ \hline
    Oxford Flowers  &89.47\scriptsize{$\pm$0.08}&   \textbf{98.04\scriptsize{$\pm$0.09}} \\ 
    Oxford-IIIT Pets   &89.73\scriptsize{$\pm$0.18} &  \textbf{92.53\scriptsize{$\pm$0.12}}  \\
    FGVC-Aircraft   &77.25\scriptsize{$\pm$0.09}&  \textbf{80.76\scriptsize{$\pm$0.12}} \\ 
    Stanford Cars  &82.13\scriptsize{$\pm$0.03}& \textbf{91.89\scriptsize{$\pm$0.07}} \\ 
        \bottomrule[1.pt]
    \end{tabular}}
    \label{tab:fine-grained}
    \hspace{5mm}

    \captionof{table}{Test accuracy (\%) on ImageNet-1k with ViT-Base/Large/Huge.}
	\renewcommand\arraystretch{.95}
	\resizebox{0.92\textwidth}{!}{
        \begin{tabular}{l|c c}\toprule[1.pt]
    Model & baseline & Ours \\ \hline
    ViT-B & 82.30 & \textbf{83.38}\scriptsize{$\uparrow$1.08} \\
    ViT-L & 84.47 & \textbf{85.01}\scriptsize{$\uparrow$0.54} \\
    ViT-H & 85.91 & \textbf{86.88}\scriptsize{$\uparrow$0.97}\\
        \bottomrule[1.pt]
    \end{tabular}}
    \label{tab:vit}
    \end{minipage}
    \hspace{3mm}
    \begin{minipage}{0.66\textwidth}
    \begin{subfigure}[]{0.5\linewidth}
		\centering
		\includegraphics[width=1.\textwidth]{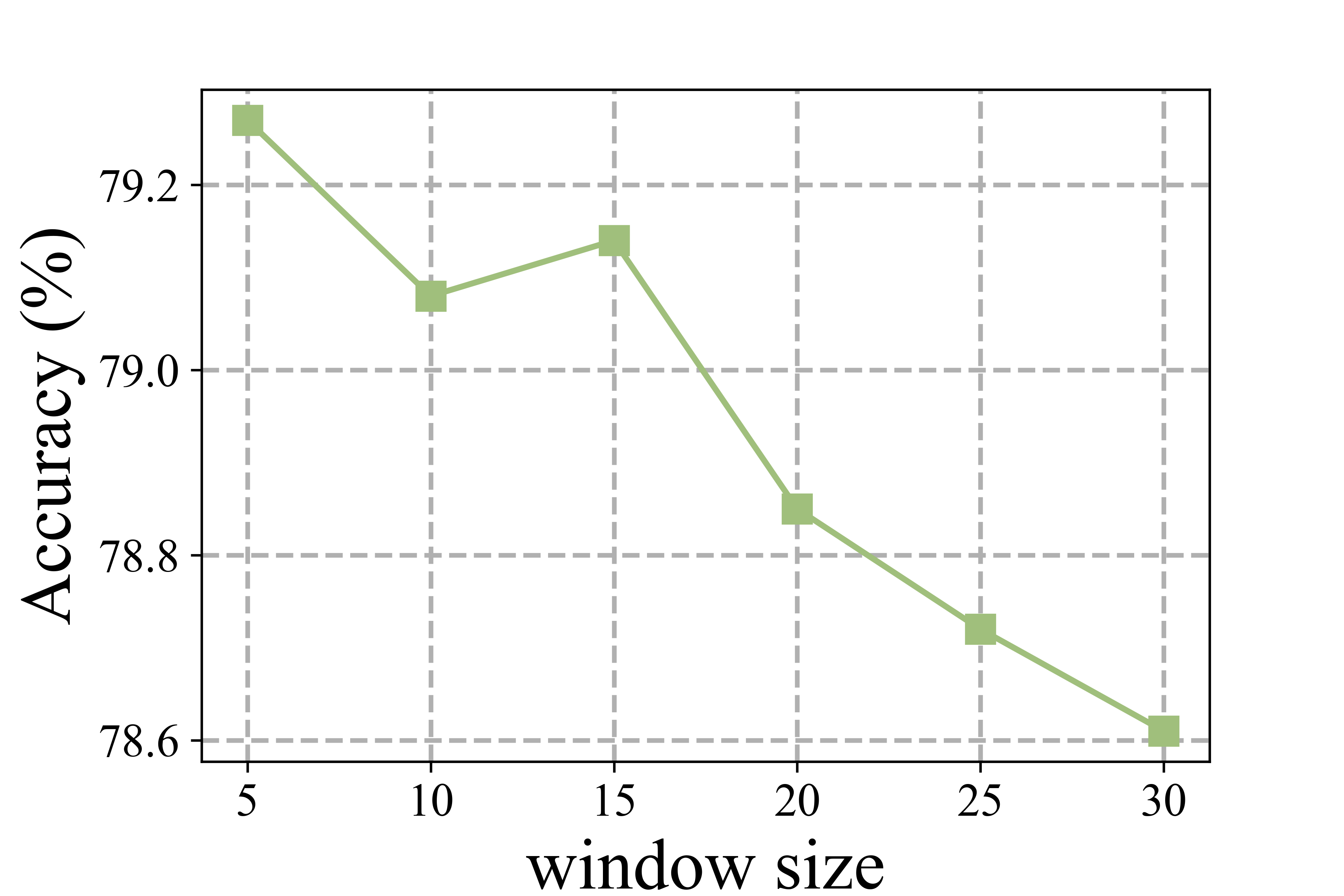}
		\caption{}\label{fig:ablation-window-size}
		\label{fig1-2}
	\end{subfigure}
	\begin{subfigure}[]{0.5\textwidth}
		\centering
		\includegraphics[width=1.\textwidth]{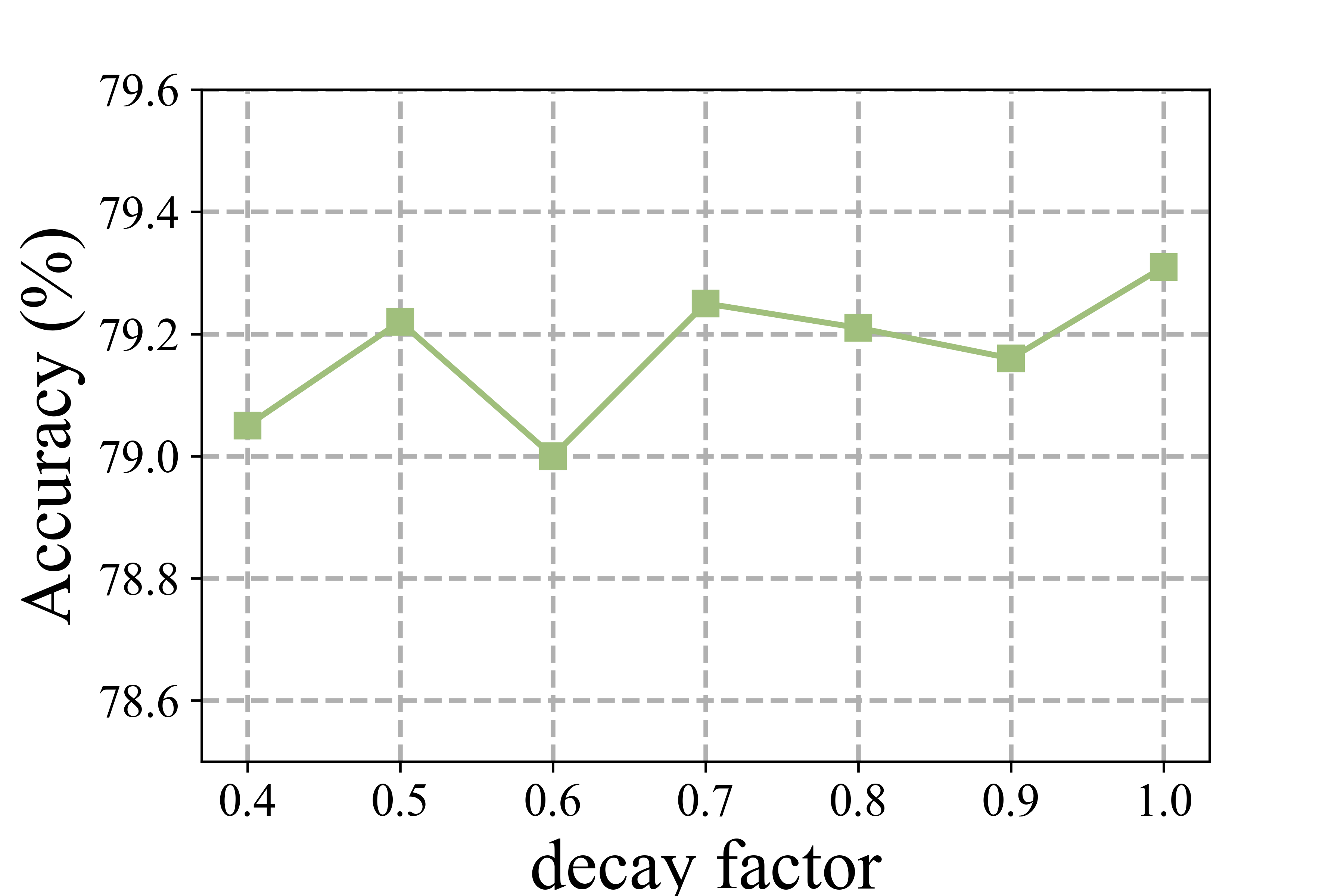}
		\caption{}\label{fig:ablation-decay}
		\label{fig1-1}
	\end{subfigure}
	\caption{The stability of our method on the two parameters, i.e., the window size and the decay factor, with CIFAR-100 using ResNet-18.}
	\label{fig:ablation-study}
    \end{minipage}
 
    \vspace{-5mm}
\end{figure*}

\subsection{Results on Fine-Grained Datasets}
To further assess the versatility of our method, we evaluate its performance on several fine-grained classification benchmarks, including Oxford Flowers~\citep{oxford-flower}, Oxford-IIIT Pets~\citep{oxford-pets}, FGVC-Aircraft~\citep{oxford-aircraft}, and Stanford Cars~\citep{stanford-cars}.
These datasets are characterized by subtle inter-class differences, making them particularly challenging for standard data augmentation strategies.

As shown in Table~\ref{tab:fine-grained}, incorporating our method into the standard training process can significantly enhance model performance.
Notably, on the Oxford Flower dataset, it achieves over 8\% absolute improvement compared to baseline learning.
These results highlight the effectiveness of our sample-aware augmentation approach in fine-grained scenarios.


\subsection{Results on Long-Tailed Datasets}
While most existing DA methods are not evaluated on long-tailed datasets, we further evaluate the robustness of our method on more challenging long-tailed benchmarks, i.e., ImageNet-LT and Places-LT~\citep{lt-datasets}, which exhibit significant class imbalance. 
We closely follow the experimental setting in OLTR~\citep{lt-datasets}, using the same network backbone and evaluation metrics, except utilizing our augmentation method.
As shown in Table~\ref{tab:imagenet-LT}, our method achieves consistent performance improvements across both closed-set and open-set evaluation settings.
On ImageNet-LT, we improve the overall top-1 accuracy by 1.3\% in the closed-set scenario.
On Places-LT, our method increases the F-measure by 4\% in the open-set setting.
These results highlight the ability of our adaptive augmentation strategy to improve generalization under severe data imbalance, without requiring explicit rebalancing techniques or auxiliary supervision.



\subsection{Cross-Architecture Generalization}
In Table~\ref{tab:cifar} and Table~\ref{tab:tiny}, we demonstrate the effectiveness of our method across various CNN-based architectures.
To further evaluate its generalizability, we extend our experiments to Vision Transformer-based models using the ImageNet-1k dataset.
As shown in Table~\ref{tab:vit}, our method yields consistent performance gains for both ViT variants, improving the performance of ViT-Base/Large/Huge on ImageNet-1k.
Importantly, these gains are achieved without introducing large additional training overheads, highlighting the efficiency of our method.
Consequently, these results confirm that our method is architecture-agnostic and can be seamlessly integrated into training pipelines as a plug-and-play module to improve performance.

\begin{wraptable}[15]{r}{0.55\linewidth}
    \vspace{-1.2cm}
    \setlength{\tabcolsep}{2pt}
  \centering
    \includegraphics[width=1\linewidth]{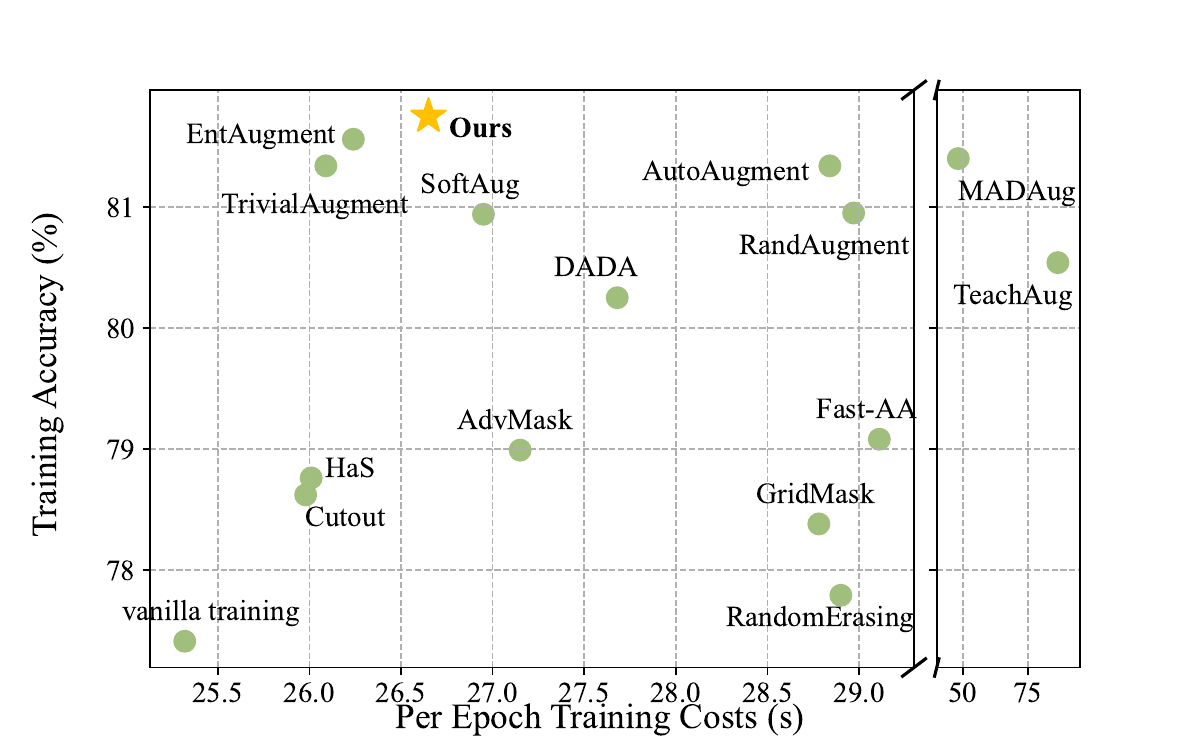}
    \captionof{figure}{Comparison in the effectiveness-efficiency tradeoff. We report the average per-epoch training costs using a 2-NVIDIA-RTX2080TI-GPUs server. }
    \label{fig:tradeoff}
\end{wraptable}
\subsection{Efficiency Comparison}
We compare the training costs of our method with other baselines.
As illustrated in Figure~\ref{fig:tradeoff}, in the efficiency-effectiveness plane, our method achieves a favorable trade-off between training cost and performance.
Consistent with the complexity analyses in Section~\ref{sec:method}, our approach introduces negligible additional overhead compared to standard training. 
This is primarily because the required gradient information can be directly obtained during standard forward and backward passes, without relying on auxiliary networks or a complex optimization process.
While our method incurs slightly higher training costs than baselines such as Cutout, HaS, and TrivialAugment, the difference is minimal.
Importantly, our method consistently delivers better performance, achieving a better balance between efficiency and accuracy.

\subsection{Ablation Study}
We conduct an ablation study to investigate the effect of two hyperparameters in our method: the window size $L$ in Eq.~\eqref{eq:window-k} and decay factor in Eq.~\eqref{eq:ema}.
As shown in Figure~\ref{fig:ablation-study}(a), increasing the window size $L$ leads to a consistent drop in classification accuracy.
This is because larger windows oversmooth the instantaneous dynamics of sample influence, thereby delaying the dynamic augmentation's responsiveness to model training dynamics.
As a result, maintaining a small window size not only better captures the local importance of each sample but also reduces the memory costs.
Figure~\ref{fig:ablation-study}(b) shows the effect of varying the decay factor $\beta$.
The model performance remains generally stable across different $\beta$ values, indicating that our method is robust to it.

\section{Conclusion}
This paper proposes a novel on-the-fly data augmentation method that performs sample-aware augmentation by modeling the evolving interplay between data and the model during training.
Unlike existing approaches, our proposed method leverages a dynamic augmentation mechanism, mitigating overfitting for stable samples by increasing their diversity while promoting generalization for uncertain ones by preserving semantic fidelity.
We hope our work inspires further research on train-dynamic-aware data augmentation from an on-the-fly perspective and believe our method will serve as a promising plug-and-play tool for the community, enabling enhanced modern deep learning training.


\bibliography{iclr2026_conference}
\bibliographystyle{iclr2026_conference}

\end{document}